\pdfoutput=1
\documentclass[11pt]{article}

\usepackage{ACL2023}

\usepackage{times}
\usepackage{color} 
\usepackage{latexsym}
\usepackage{subcaption}

\usepackage{amsmath,mathtools}
\usepackage{color,graphicx}
\usepackage{pgfplots}
\usepackage{adjustbox}
\usepackage{enumitem}

\usepackage{tikz}
\usepackage{tikz-dependency}
\usetikzlibrary{matrix}
\usetikzlibrary{%
  shapes,%
  arrows,%
  positioning,%
  calc,%
  automata,%
  chains,%
  fit%
}
\usepackage{dblfloatfix}

\usepackage{titlesec}
\titleformat{\paragraph}
{\normalfont\normalsize\bfseries}{\theparagraph}{1em}{}
\titlespacing*{\paragraph}
{0pt}{3.25ex plus 1ex minus .2ex}{1.5ex plus .2ex}

\definecolor{pf7}{RGB}{166, 118, 29}
\definecolor{myblue}{RGB}{80,80,160}
\definecolor{mygreen}{RGB}{80,160,80}

\usepackage[T1]{fontenc}
\usepackage[utf8]{inputenc}
\DeclareUnicodeCharacter{2212}{-}

\usepackage{CJKutf8}

\usepackage{microtype}
\usepackage{inconsolata}

\newcommand{\citenopar}[1]{\citeauthor{#1}, \citeyear{#1}}

\newcommand\omicron{o}

\title{Assessing the Role of Lexical Semantics in Cross-lingual Transfer through Controlled Manipulations}

\author{Roy Ilani, Taelin Karidi, Omri Abend\\ 
Hebrew University of Jerusalem \\\texttt{\small roy54x@gmail.com, \{taelin.karidi,omri.abend\}@mail.huji.ac.il}}

\begin{document}

\begin{CJK*}{UTF8}{gbsn}

\maketitle

\begin{abstract}
While cross-linguistic model transfer is effective in many settings, there is still limited understanding of the conditions under which it works. In this paper, we focus on assessing the role of lexical semantics in cross-lingual transfer, as we compare its impact to that of other language properties. Examining each language property individually, we systematically analyze how differences between English and a target language influence the capacity to align the language with an English pretrained representation space.
We do so by artificially manipulating the English sentences in ways that mimic specific characteristics of the target language, and reporting the effect of each manipulation on the quality of alignment with the representation space. We show that while properties such as the script or word order only have a limited impact on alignment quality, the degree of lexical matching between the two languages, which we define using a measure of translation entropy, greatly affects it.\footnote{Our code is available at \url{https://github.com/roy54x/Lexical_semantics_in_cross-lingual_transfer}.}
\end{abstract}

\section{Introduction}

\begin{figure}[h]
    \includegraphics[width=1.0\linewidth]{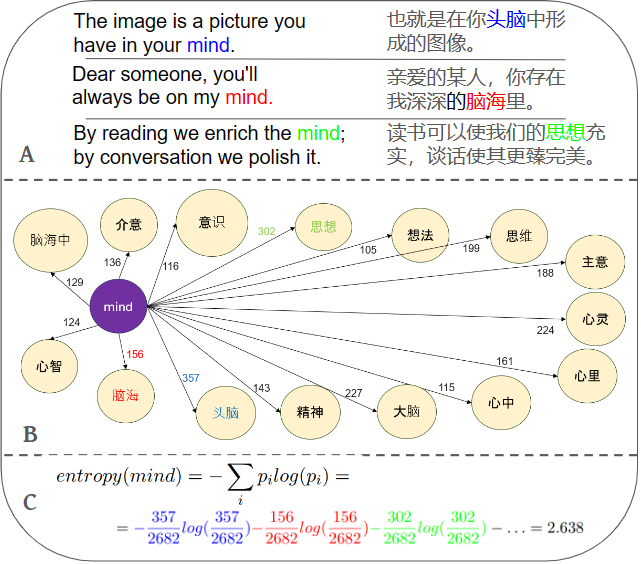}
    \caption{\textbf{A.} Sentences from the \href{http://nlp2ct.cis.umac.mo/um-corpus/}{\textit{UM}} parallel corpus. In each sentence, the word \textit{mind} is colored along with its translation in Simplified Chinese. \textbf{B}. A weighted graph which results from the UM corpus. The edge weights indicate how many times \textit{mind} is translated into each instance in Simplified Chinese. \textbf{C.} Calculation of the \textit{translation entropy} of the word \textit{mind} in the UM corpus.}
    \label{fig:mind-entropy-chinese}
\end{figure}

Different languages partition meanings over their vocabularies in different ways. In English, the concept \textit{wall} includes both a structural component in a house and a defensive barrier around a city, whereas Spanish distinguishes them with the concepts \textit{pared} and \textit{muro}. This raises the question of how such differences in lexical semantics influence cross-lingual transfer -- the ability of models trained on data from one language to effectively perform tasks in another language \citep{kim-etal-2017-cross,artetxe-schwenk-2019-massively, Dobler_2023}.

%While multilingual NLP is gaining increasing attention, a performance gap persists between English and other languages, particularly low-resource ones. To address this, various cross-lingual transfer techniques have been proposed , including training shared multilingual representation spaces \cite{artetxe-etal-2018-robust,ruder2019survey, Heffernan:2022, tan-etal-2023-multilingual}.

In this work, we study the impact of lexical semantics and other linguistic properties on the effectiveness of cross-lingual transfer. We examine how various properties affect the ability to extend an existing representation space to include an additional low-resource language, and consequently, how they affect the zero-shot performance of the low-resource language. 

To isolate the distinct linguistic properties and evaluate their individual impact, we perform manipulations to the English language that mimic specific language traits found in other languages, thereby creating artificial languages. For instance, to evaluate the impact of lexical semantics, we create an artificial language by imposing lexicalization patterns of other languages onto English.

We define a weighted bipartite graph that links the vocabularies of two languages, mapping each word in one language to all its potential translations in the other language. We leverage this graph to characterize the lexicalization patterns between the languages in information theoretic terms.

Our results indicate that the lexicalization patterns of the source and target languages have more impact on transferability than other linguistic properties. They also demonstrate robust correlation between the entropy of words in the bipartite graph we define and zero-shot performance. 

%%%%%%%%%%%%%%%%%%%%%%%%%%%%%%%%%%%%%%%%%%%%%

\section{Related Work}

\subsection{Cross-lingual Transfer Methods}

Multilingual language models (MLLMs) like mBERT \citep{Devlin:2019} and XLM-R \citep{Conneau:2020a} exhibit remarkable zero-shot cross-lingual performance, despite being trained without parallel data. However, they also face limitations. Being contextualized token embeddings, they may underperform in sentence-level tasks \citep{Hu:2020b}. Moreover, training these models requires a massive amount of text from each language, posing a major challenge to the inclusion of low-resource languages.

To overcome these limitations, \citet{Reimers:2019, Reimers:2020} trained a model (SentenceBERT) using a sentence-level objective to obtain sentence representations \citeyearpar{Reimers:2019}. They then employed knowledge distillation (teacher-student supervised learning) to extend the representation space to additional languages \citeyearpar{Reimers:2020}. This approach, while requiring parallel data, proves effective with relatively few samples, making it suitable for low-resource languages. \citet{Heffernan:2022} applied a similar technique with LASER2 \citep{Artetxe:2019b}, a language-agnostic sentence encoder, as their teacher model. They demonstrated the efficiency of this approach with extremely low-resource languages. In our research, we follow a similar setup.

\subsection{Investigations of Zero-shot Transfer}

Many studies examined the factors affecting zero-shot cross-lingual transfer. Some defined metrics of language similarity, such as geographical, genetic, or phonological distance, and explored their relation to transferability \cite{lin-etal-2019, lauscher-etal-2020-zero, Dolicki-el-al, ahuja-etal-2022-multi}. Others focused on assessing the impact of specific properties like lexical overlap \citep{wu-dredze-2019-beto, patil-etal-2022-overlap, de-vries-etal-2022-make} or syntactic structure and word order \citep{dufter-schutze-2020-identifying, Arviv:2021, de-vries-etal-2022-make, Chai:2022, wu2024oolong}.

Among the studies exploring word order, some rearranged English sentences to create artificial languages and analyzed their behavior \cite{dufter-schutze-2020-identifying, deshpande-etal-2022-bert, Chai:2022, wu2024oolong}. As summarized by \citet{philippy-etal-2023-towards}, experiments involving sentence inversion or random shuffle showed a significant decline in zero-shot performance compared to experiments where the word order was systematically modified based on the structure of other natural languages. This implies that while word order is important, differences in word order between languages may play a minor role in zero-shot transfer.

To our knowledge, no previous work examined the impact of variations in the lexicalization patterns of the two languages on cross-lingual transfer learning. In this paper, we aim to address this gap.
In addition, while previous studies mostly focused on representations derived from masked language models, we focus on a knowledge distillation setup.

%Among the studies that investigate the impact of word order, some manipulate the English language to create artificial languages and explore their behavior \cite{dufter-schutze-2020-identifying, deshpande-etal-2022-bert, Chai:2022, wu2024oolong}.

%Many studies have been conducted to explore the factors that impact zero-shot cross-lingual transfer.\citet{Cotterell:2017} showed that the zero-shot performs better when applied to languages within the same language family. \citet{Arviv:2021} demonstrated a correlation between the preservation of syntactic relations in translation and zero-shot performance. 

%\citet{Chai:2022} conducted a similar experiment but under controlled laboratory conditions. In order to evaluate the impact of specific language properties, they create an artificial language by modifying English and pretrain a bilingual MLM with English and its modified version. Their focus was on syntactic properties, particularly word order. They observed a negligible impact on zero-shot performance when examining subtle modifications in constituent order, but a significant impact when shuffling the entire sentence randomly.

%While previous experiments focus on syntactic features, they do not address the semantic aspect. In this paper, we focus on lexical semantics.

\subsection{Lexicalization Patterns} 

Lexicalization patterns were widely used in linguistic typology to classify languages and explore language universals, in cognitive science to study conceptualisation, and even by anthropologists to examine cultural influences on language and cognition \cite{franccois2008semantic,jackson2019emotion,xu2020conceptual,georgakopoulos2022universal}. 
The majority of research on lexicalization has been centered around the concept of colexification (a linguistic phenomenon that occurs when multiple concepts are expressed in a language with the same word). Traditionally, colexification data relied on hand-curated resources, but this changed with the introduction of CLICS \cite{List:2018b}, promoting exploration into large-scale colexification graphs also in NLP \cite{harvill-etal-2022-syn2vec,liu2023crosslingual,Liu:2023}.

\citet{Liu:2023} proposed a more systematic way to investigate the conceptual relation between languages and extract colexifications. Their method includes aligning concepts in a parallel corpus and extracting a bipartite directed graph for each language pair, mapping source language concepts to sets of target language strings. Leveraging these bipartite graphs, they identify colexifications across a diverse set of languages. Here, we employ a similar method, albeit to a different purpose -- our primary focus lies in proposing a methology for quantifying how differences in lexical semantics impact cross-lingual transfer.

%%%%%%%%%%%%%%%%%%%%%%%%%%%%%%%%%%%%%%%%%%%%%%
\section{Method} 

Our main goal in this paper is to study how different language properties, with a particular emphasis on lexical patterns, influence the ability to perform cross-lingual transfer, and we aim to do so in a carefully controlled way. 

To isolate distinct language properties and understand their respective contribution, we define different manipulations of the source language $L_s$. For each of these manipulations, we modify $L_s$ so that it imitates certain properties found in a target language $L_t$, creating a new artificial language $L_{\mathcal{A}}$ (Section \S\ref{sec:manipulatios}). Throughout this article, we maintain English as the source language.

Then, for each artificial language $L_{\mathcal{A}}$, we follow the distillation method proposed by \citet{Reimers:2020}, training a model to encode sentences of $L_{\mathcal{A}}$ into an English pretrained representation space. We explore which of them allows for an effective knowledge transfer and, hence, performs well in a zero-shot setting (Section \S\ref{sec:experiments}).

\vspace{.1cm}\noindent\textbf{Model Distillation.}\label{par:knowledge-distillation} The pretrained teacher model we use is an English sentence transformer model \citep{Reimers:2019}. It is trained using English sentence pairs and a self-supervised contrastive learning objective to encode similar English sentences into vectors that are close to one another in the vector space. Given a sentence from a pair, the model is trained to predict which  of a batch of randomly sampled other sentences is in fact paired with it. The outcome of this training yields a sentence representation space that captures the semantic information of a given sentence. Within this pretrained vector space, the cosine similarity between two vectors indicates the degree of similarity between the two sentences they represent.

The pretrained representations of the teacher model serve us throughout the experiment as ground truth. For each artificial language $L_{\mathcal{A}}$, we train smaller transformer models using an \textit{English}--$L_{\mathcal{A}}$ parallel corpus. Denoting  the teacher model with $M$ and the student model that corresponds to the language $L_{\mathcal{A}}$ with $m_{\mathcal{A}}$, for each sentence pair $(s, t)\in$ \textit{English}$\times L_{\mathcal{A}}$, the training objective is to minimize the following cosine embedding loss: %cite \citet{Salvador:2017}? or maybe \href{https://pytorch.org/docs/stable/generated/torch.nn.CosineEmbeddingLoss.html}{pytorch}
\begin{equation}\small
L_{cos}(m_{\mathcal{A}}(t),M(s)) =
\begin{dcases}
        \mathbf{1-{cos}(m_{\mathcal{A}}(t),M(s))} \\
        \qquad\qquad\text{if $t$ is a manipulation of $s$}\\
        \mathbf{max(0,{cos}(m_{\mathcal{A}}(t),M(s)))}\\
        \qquad\qquad\text{otherwise}
\end{dcases}
\end{equation}
This optimization process aims to increase the cosine similarity in the vector space whenever the sentence $t$ is a manipulation of the sentence $s$, and decrease it in any other case. As a result, it produces a sentence encoder that maps each sentence $t\in  L_{\mathcal{A}}$ to a location in the pretrained vector space as close as possible to the representation of the original English sentence.

\vspace{.1cm}\noindent\textbf{Evaluation}.\label{par:evaluation} For each student model $m_{\mathcal{A}}$, we compute the average cosine similarity between the embeddings of English sentences and the embeddings of the corresponding manipulated sentences within a held-out subset of the corpus. This serves as the intrinsic evaluation. Additionally, we employ the model in a zero-shot experiment for an extrinsic NLP task and present its performance. These two outcomes help us understand the quality of the alignment of the language $L_{\mathcal{A}}$ with the pretrained representation space of the teacher model. 

%%%%%%%%%%%%%%%%%%%%%%%%%%%%%%%%%%%%%%%%%%%%%%
\section{Manipulations of the Data}\label{sec:manipulatios}

We proceed to define the different manipulations that we apply. For each manipulation, we modify the English source to create an artificial language $L_{\mathcal{A}}$, generate an \textit{English}--$L_{\mathcal{A}}$ parallel corpus, and train student models $m_{\mathcal{A}}$ as explained.

Our primary focus lies within the domain of lexical semantics. To thoroughly examine their influence, we take a broader approach, investigating how the effect of lexical semantic manipulations compare with that of other linguistic properties. We examine three aspects: script, syntax, and lexical semantics. For each of these aspects, we define a manipulation that solely modifies it. First, we substitute the letters of the English alphabet with symbols of a different script (\S\ref{ssec:script}). Second, we systematically rearrange the word order in sentences, thus examining the effect of the syntactic structure, or at least a specific aspect of it (\S\ref{ssec:word_order}). Finally, we replace the English lexicon with that of a target language to explore the significance of variations in lexicalization patterns (\S\ref{sec:method-lexical}). By isolating each linguistic aspect, we obtain a clearer understanding of its individual contribution.

\subsection{Manipulating the Script}\label{ssec:script} 

To manipulate the script we simply substitute each English character with a symbol from another script in an injective manner. For instance, if we consider the Greek alphabet system, we can swap the characters according to their sequential order: $a\rightarrow \alpha$, $b\rightarrow \beta$, $c\rightarrow \gamma$, and so forth. This way, the sentence \textit{Brown cows eat grass} will transform into: $\beta \sigma \omicron \psi \xi$ $ \gamma \omicron \psi \tau$ $\epsilon \alpha \upsilon$ $\eta \sigma \alpha \tau \tau$.

\subsection{Manipulating Word Order}\label{ssec:word_order}

The second manipulation we use is a word reordering one. We apply the word reordering algorithm developed by \citet{Arviv:2023}, to permute the words of each source sentences so that it will conform to the syntactic structure of the target language (see Appendix \ref{app:reorder-algorithm} for full details).
The algorithm recursively reorders all the subsequences in a source sentence, yielding a new sentence in an artificial language $L_{\mathcal{A}}$ that imitates the word-order of a target language $L_t$.
For example, the sentence \textit{Brown cows eat grass} yields different results depending on the selected target language. Spanish, an \textit{SVO} language, but in which nouns are ordered before adjectives, produces: \textit{Cows brown eat grass}; whereas Hindi, an \textit{SOV} language, produces: \textit{Brown cows grass eat}.\footnote{Since the algorithm is based on fixed statistics, the artificial language it produces exhibits a more consistent word order than that of a natural language. We prefer this experiment over one in which the order of words in a sentence is randomly rearranged due to the potential noise this might add.}

\subsection{Manipulating the Lexicon}\label{sec:method-lexical} 

The core of our study is lexical semantics and their impact on cross-lingual transfer. We seek to assess the influence of the diverse distribution of meanings across different lexicons. To achieve this, we develop a manipulation in which we substitute the lexicon of the source language $L_s$ with that of a target language $L_t$. This creates a new artificial language $L_{\mathcal{A}}$ that is based on the lexicon of $L_t$ while retaining the original sentence structure of $L_s$.

The manipulation is based on a word alignment between a source sentence $s\in L_s$ and its translation in the target language $t\in L_t$. We replace each word in the source language with its corresponding translation in the target language, thus adopting the lexical semantics of the target language while preserving the original syntax.\footnote{This manipulation inherently includes the first manipulation, at least to some extent, as altering specific words in the sentence also influences the script. However, we will demonstrate later that the script is not a significant factor, making this fact of minor importance to our conclusions.}

However, word-aligned bitext is difficult to obtain. While manually aligned parallel datasets are scarce and limited in size, model-based automatic aligners often map words to a wide range of possible translations. When defining a lexical manipulation that depends on mapping one lexicon to another, ensuring each word consistently maps to the same set of words is crucial. To address this, we develop an algorithm that refines the output of an automatic aligner, mapping each word of the lexicon to a fixed set of words. This careful process involves extracting a bipartite graph from the bitext, which ensures consistent mapping.

\vspace{.1cm}
\noindent\textbf{Formalism.} Consider a word-aligned bitext that contains the languages $L_s$ and $L_t$. We define $G=(V_s,V_t,E,w)$ to be a weighted bipartite graph, where $V_s$ is the set of words in the lexicon of $L_s$, and $V_t$ is the set of words in the lexicon of $L_t$. A pair of words $(v,u)\in V_s\times V_t$ is an edge in $G$ iff $v$ is aligned to $u$ in at least one instance in the bitext. The weight function $w:E\rightarrow N^+$ assigns the number of times that each word pair is aligned in the bitext.

\begin{figure}
\begin{tikzpicture}[node distance={5mm}, main/.style = {draw, circle}, every fit/.style={ellipse,draw,inner sep=-2pt,text width=2cm}]

\begin{scope}[start chain=going below,node distance=5mm]
\node[main] (1) {\textit{for}}; 
\node[main] (2) [below=1.0] {\textit{by}}; 
\end{scope}

\begin{scope}[xshift=5cm,yshift=0cm,start chain=going below,node distance=5mm]
\node[main] (3) {\textit{por}};
\node[main] (4) [below=1.0] {\textit{para}};
\end{scope}

\node [fit=(1) (2),label=above:\textit{English}] {};
\node [fit=(3) (4),label=above:\textit{Spanish}] {};

\draw (1) -- node[midway, above, sloped]{85303} (3);
\draw (1) -- node[midway, above left, sloped]{175771} (4);
\draw (2) -- node[midway, below left, sloped]{93781} (3);

\end{tikzpicture} 
\caption{Illustration of the weighted sub-graph which results from the \href{https://www.statmt.org/europarl/}{\textit{Europarl}} parallel corpus. The edges represent the possibility that two words are translations of each other. The weights denote the number of occurrences that each word pair is aligned in the bitext.}\label{fig:weighted_subgraph}
\end{figure}
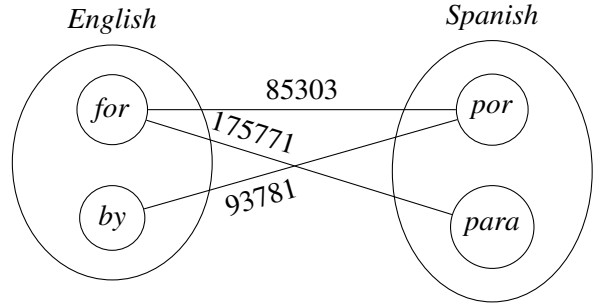

This construction aims to capture the relationship between the lexical semantics of two languages. For example, the Spanish translation of \textit{for} is \textit{por} in some cases and \textit{para} in others; \textit{by} is also occasionally translated as \textit{por} (e.g., \textit{multiply by three} translates to \textit{multiplicar por tres}). We thereby obtain the subgraph in Figure \ref{fig:weighted_subgraph}.

We hypothesize a negative correlation between the degree of the vertices in the graph and the ability to perform cross-lingual transfer between the languages. In other words, the closer the lexicons approximate a bijective relationship, the better we expect the cross-lingual transfer performance to be.

\vspace{.3cm}\noindent\textbf{Swapping Algorithm.} 
We proceed to outline the systematic procedure we employ to perform the lexical manipulation. For each pair of languages $L_s$, $L_t$ we follow these steps:

\begin{enumerate}[wide, labelwidth=!, labelindent=0pt, itemsep=3pt]
\item 
We apply an automatic word aligner %(see details in Section \S\ref{sec:Models-details}) 
to a large $L_s$--$L_t$ parallel corpus, extracting a weighted bipartite graph $G$ as described above.

\item 
We filter out of the graph any edge $e\in E$ that represents an alignment which is not substantial (that is, an edge whose weight does not exceed a certain threshold or whose weight is relatively small compared to other edges originating from the same vertex).\footnote{These parameters may depend on the target language. See Appendix \ref{par:graph-filtering}.}

\item Given a source sentence $s\in L_s$ and its translation in the target language $t\in L_t$, we run the automatic aligner to achieve a word-to-word alignment between $s$ and $t$.

\item For each source word $v\in s$:

\begin{enumerate}
\item If there exists a word $u\in t$ such that the word-to-word alignment includes the pair $(v,u)$, and at the same time it holds that $(v,u)\in E$, then we replace the word $v$ with $u$.

\item Otherwise, if there exists a word $u\in t$ such that $(v,u)\in E$, we replace the words as well. If there is more than one valid choice, we select the word $u\in t$ for which the weight $w(v,u)$ is the highest.

\item\label{item:most-common-word} Otherwise, we look for the edge $(v,u)\in E$ that has the highest weight among all edges originating from $v$, meaning that $u$ is the most common alignment of $v$ in the language $L_t$. 
If we find such an edge, we replace $v$ with $u$.\footnote{In languages where the words have different inflections, we check the validity of the match based on the lemma, but replace the words in their original form. The determination of the most common alignment also considers the original inflection.}

\item In a case there are no edges originating from $v$, we preserve it. 

\end{enumerate}
\end{enumerate}

This systematic procedure provides a mapping between two lexicons, and therefore enables us to make consistent decisions for each word in the lexicon – whether to be replaced or preserved. This helps maintaining a coherent semantic structure in the resulting artificial language $L_{\mathcal{A}}$.\footnote{To ensure that our algorithm does not harm the alignments of the auto-aligner, we compare the precision and recall of our algorithm's outputs with the original alignments of the auto-aligner against the gold standard. We observe higher precision but lower recall, resulting in a slightly better F1 score for our algorithm's outputs. For further details, please refer to Appendix \ref{app:golden-alignments}.}

For a simple illustration, consider the sentence \textit{Brown cows eat grass} and its Spanish translation \textit{Las vacas marrones comen hierba}, and assume the auto-aligner's output is \textit{brown}$\rightarrow$\textit{marrones}, \textit{eat}$\rightarrow$\textit{comen}. When applying our swapping algorithm, we first check whether the edges \textit{(brown--marr$\Acute{o}$n)} and \textit{(eat--comer)}\footnote{The lemmas of \textit{marrones} and \textit{comen} respectively.} appear in the bipartite graph. As both of them do, we replace \textit{brown} with \textit{marrones} and \textit{eat} with \textit{comen}. Next, we search for words in the target sentence that are linked to the source words in the graph, resulting in the edges \textit{(cow--vaca)} and \textit{(grass--hierba)}. These two words are swapped with their corresponding pairs as well. This process ultimately yields the sentence: \textit{Marrones vacas comen hierba}.

\vspace{.2cm}\noindent\textbf{Translation Entropy.} 
To further appreciate the impact of the divergence between the source and the target lexicons, we introduce the concept of \textit{translation entropy}. Let $G$ be the weighted bipartite graph presented earlier, we compute the entropy for each vertex $v$ in the graph:\footnote{It does not matter whether $v\in L_s$ or $v\in L_t$.}
\begin{equation}
e(v) = -\sum_{u\in U_{v}}p_{v}(u)log(p_{v}(u))
\label{eq:entropy}
\end{equation}
where $U_{v}$ is the subset of vertices linked to $v$, and $p_{v}$ is the following probability function:
\begin{equation}
p_{v}(u)=\frac{w(v,u)}{\sum_{u^\prime\in U_{v}}w(v,u^\prime)}
\label{eq:probability-function}
\end{equation}
As $w$ counts occurrences of each word pair aligned in the bitext, the outcome of the function $p_{v}$ is the probability that, in a particular instance, the word $v$ is linked to the word $u$ among all possible $u^\prime\in U_{v}$ (see calculation example in Figure \ref{fig:mind-entropy-chinese}).

We examine the impact of \textit{translation entropy} in two distinct configurations: one for the source words (Figure \ref{fig:src-trg-entropy-filtering}{A}) and another for the target words (Figure \ref{fig:src-trg-entropy-filtering}{B}). In the first, we compute the \textit{translation entropy} for all source vertices $v\in V_s$ and partition the set $V_s$ into three disjoint subsets based on the percentile of the \textit{translation entropy} values. The percentile calculation is based on the number of instances in the database, rather than the number of words in the lexicon. Then, in each experiment, we remove from the graph $G$ all the source vertices that do not belong to a specific subset to achieve the sub-graph $G^\prime$. We apply once again the lexical manipulation, but this time using the filtered graph $G^\prime$. In the second configuration, we follow the exact same steps, but this time for the target vertices $v\in V_t$.\footnote{The two configurations are not directly comparable: removing a source word from the lexicon leads to a reduction in the number of swaps performed, whereas removing a target word reduces the diversity of the swaps but not necessarily the number of them.}

Returning to the previous example, after we extract an \textit{English--Spanish} bipartite graph from the \href{https://www.statmt.org/europarl/}{\textit{Europarl}} parallel corpus and compute the entropy of the English words, we obtain: \textit{e(brown)=0.545}, \textit{e(cow)=0.24}, \textit{e(eat)=0.631}, \textit{e(grass)=0.799}. If we filter the graph to retain only words that fall within the upper third of \textit{translation entropy} values, we find that the word \textit{grass} is the only one meeting this criterion. Consequently, applying our lexical manipulation to this filtered graph results in the sentence: \textit{Brown cows eat hierba}.

%%%%%%%%%%%%%%%%%%%%%%%%%%%%%%%%%%%%%%%%%%%%%

\section{Experimental Setup} 

\subsection{Datasets}

In this subsection, we outline the datasets used in our work. For further details on the datasets, their selection rationale, and the filtering process applied, please refer to Appendix \ref{app:datasets}. The bitext we use for training and intrinsic evaluation of the student models throughout all manipulation experiments is the \href{https://opus.nlpl.eu/TED2020.php}{\textit{TED-2020}} parallel corpus \citep{Reimers:2020}. To avoid over-specializing the tokenizer on this small dataset, we train the tokenizers on the \href{https://huggingface.co/datasets/cc100}{\textit{CC-100}} corpus \citep{Weznek:2020}.\footnote{Full details on tokenizer training for each artificial language can be found in Appendix \ref{app:tokenizers}.} The CC-100 corpus is also used for our monolingual benchmark experiments (see beggining of Section \S\ref{sec:experiments}). For extrinsic evaluation we use the \href{https://github.com/facebookresearch/XNLI}{\textit{Cross-lingual Natural Language Inference (XNLI; \citenopar{conneau:2018})}}. To extract the bipartite graphs, we require a large parallal corpus. Therefore, we use the \href{https://www.statmt.org/europarl/}{\textit{Europarl}} corpus for European languages and the \href{http://nlp2ct.cis.umac.mo/um-corpus/}{\textit{UM}} corpus \citep{Tian:2014} for Simplified Chinese.

\subsection{Models}\label{sec:Models-details}

\textbf{Teacher Model}. We select the pretrained sentence transformer \href{https://huggingface.co/sentence-transformers/all-mpnet-base-v2}{\textit all-mpnet-base-v2}. This model was trained on 1B English sentence pairs with a self-supervised contrastive learning objective (see Section \S\ref{par:knowledge-distillation}). The training produced a 768-dimensional vector space that has proven to achieve state-of-the-art results in sentence-level tasks.

\vspace{.1cm}\noindent\textbf{Student Models}. We train multiple RoBERTa models \cite{Liu:2019}, with each model designed to encode a sentence into the teachers' 768-dimensional vector space. To achieve this, we add a mean-pooling layer on top of the last hidden layer. We set the vocabulary size to 30527, matching that of the teacher model, the number of max position embeddings to 28, and the hidden size to 768. As to the number of hidden layers and the number of attention heads, we explore various architectures: 3/6/9/12 hidden layers paired with 4/6/8/12 attention heads, respectively. The number of trainable parameters for these configurations are 84316, 135004, 185692, and 236380, respectively. We reserve a small portion of the dataset for testing (20K sentence pairs in the \textit{TED} corpus and 100K sentences in \textit{CC-100}), and then randomly split the training set into 90\% for actual training and 10\% for validation. We use the Adam optimizer with a learning rate of $3e^{-5}$, continuing until the validation loss does not decrease for five consecutive epochs. The model with the lowest validation loss is selected, and its performance on the test set is reported.

\vspace{.1cm}\noindent\textbf{NLI Model}. For the zero-shot English NLI experiment, we train a Multi-Layer Perceptron (MLP) on top of the teacher model. We use the usual combination of the two sentence embeddings: ($p$; $h$; $p\cdot h$; $|p-h|$), where $p$ and $h$ are the premise and the hypothesis respectively (see for example \citenopar{Artetxe:2019b}). We build the MLP with two hidden layers of size 128, and train it for 150 epochs using the Adam optimizer. We select the model that achieves the lowest loss on the test set.

\vspace{.1cm}\noindent\textbf{Auto-aligner}. For obtaining high-quality word-to-word alignments we use the \href{https://github.com/cisnlp/simalign}{\textit{Simalign}} automatic aligner \cite{Jalili:2020}. This tool uses contextualized embeddings to map words between sentences. We run it with \textit{XLM-R} as the base model, and set the matching method to be $ArgMax$. To simplify the analysis, we filter its outputs to include only one-to-one alignments. We manually review some of the alignments to ensure their quality.

\section{Experiments $\&$ Results}\label{sec:experiments}

To assess the impact of each linguistic property on transferability, we apply our manipulations to English and carry out the distillation process for each artificial language $L_{\mathcal{A}}$. For intrinsic evaluation, we compute the average cosine similarity (hereafter: ACS) between the embeddings of English sentences and the embeddings of the corresponding manipulated sentences in the test set. For extrinsic evaluation, we use XNLI zero-shot accuracy.

Before manipulating English, we conduct experiments to obtain reference points for evaluating the models. First, we perform the distillation process on regular English sentences from the CC100 corpus. We explore the influence of varying training set sizes and of the student model architecture. We observe a considerable margin, with differences of up to 0.227 in ACS, between models trained on 50K sentences and those trained on 1M. Conversely, we observe a smaller margin, with differences of up to 0.035 in ACS, between smaller and larger model architectures. Our findings suggest that for low-resource scenarios, exceeding 6 hidden layers and 6 attention heads is unnecessary. For full results refer to Appendix \ref{app:english-english}.

Second, we conduct cross-lingual experiments using the \href{https://opus.nlpl.eu/TED2020.php}{\textit{TED-2020}} parallel corpus to compare English with other natural languages without manipulation. The results of the cross-lingual experiments serve as a lower bound for the performance on the manipulated data (as the manipulations are meant to change English to be closer to the target language). We also train student models on English with a newly trained tokenizer to arrive at an upper bound. We observe a substantial range between the lower and upper bounds, with differences of up to 0.186 in ACS, which gives us sufficient room to experiment with our manipulations. See full results in Appendix \ref{app:cross-lingual}.

We proceed to apply our manipulations to tease apart the properties of the data that contribute to this difference between in-language training and cross-lingual training.

\vspace{-.2cm}
\subsection{Script Substitution}\label{sec:script}

We perform two script substitutions: first, replacing the English characters with Greek characters sorted alphabetically, and second, replacing them with Simplified Chinese characters sorted by their frequency. For all the student models, we train a tokenizer from scratch (see Appendix \ref{app:tokenizers} for full details). The results, reported in Table \ref{tab:script-substitution}, show almost no degradation in performance.\vspace{-.2cm}

%We perform two script substitutions, replacing the English characters first with Greek characters, sorted alphabetically, and second with Simplified Chinese characters, sorted by their frequency. The results, reported in Table \ref{tab:script-substitution}, show no degradation in performance when compared to the encoders trained with a new tokenizer (see Appendix \ref{app:tokenizers}). This suggests that substituting the script has no effect as long as a tokenizer is trained on the new script .

%\begin{table}[h!]
%    \centering
%    \small{
%    \begin{tabular}{|p{2.5cm}||p{0.7cm}|p{0.7cm}||p{0.7cm}|p{0.7cm}||}
%    \hline
%    \space & \multicolumn{2}{l||}{\mbox{Similarity score}} & \multicolumn{2}{l||}{\mbox{XNLI accuracy}} \\[0.5ex]\cline{2-5}
%     \space &\textbf{50K} & \textbf{100K} & \textbf{50K} & \textbf{100K} \\ 
%    \hline\hline
%    English (new tok.) & 0.725 &  0.786 &  55.7 & 59.4\\[0.5ex] \hline
%    Greek alphabet & 0.725 & 0.786 & 54.6 & 58.7 \\[0.5ex] \hline
%    Chinese symbols & 0.728 & 0.788 & 55.3 & 58.4 \\[0.5ex] \hline
%    \end{tabular}
%    }

%    \caption{Results from the distillation process for the script substitution experiment.}\label{tab:script-substitution}
%\end{table}

\begin{table}[h!]
    \centering
    \small{
    \begin{tabular}{|p{2.3cm}||p{0.75cm}|p{0.75cm}||p{0.75cm}|p{0.75cm}||}
    \hline
    \space & \multicolumn{2}{c||}{ACS score} & \multicolumn{2}{c||}{XNLI accuracy} \\[0.5ex]\cline{2-5}
     \space &\textbf{50K} & \textbf{100K} & \textbf{50K} & \textbf{100K} \\ 
    \hline\hline
    English & 0.725 & 0.786 & 55.7 & 59.4\\[0.5ex] \hline
    Greek alphabet & -0.0 & -0.0 & -1.1 & -0.7 \\[0.5ex] \hline
    Chinese symbols & +0.003 & +0.002 & -0.4 & -1.0 \\[0.5ex] \hline
    \end{tabular}
    }
    \caption{Results for the script substitution experiment. 50K/100K denote for the number of training sentences.\vspace{-.1cm}}\label{tab:script-substitution}
\end{table}

\vspace{-.2cm}
\subsection{Word Reordering}\label{sec:word-reorder}

We apply the reordering algorithm developed by \citet{Arviv:2023} each time relying on the \emph{pairwise ordering distributions} of a different language. We examine \textit{SVO} languages (Spanish, Greek, Chinese and Hebrew) as well as an \textit{SOV} language (Hindi). Results are presented in Table \ref{tab:word-reorder}. Although we observe a degradation in performance, it is a very slight one. The ACS score in the worst case (100K Greek sentences) decreases by 0.013 points, and the XNLI accuracy in the worst case (100K Hindi sentences) decreases by 1.5\%. 

%\begin{table}[h!]
%    \centering
%    \small{
%    \begin{tabular}{|p{2.5cm}||p{0.7cm}|p{0.7cm}||p{0.7cm}|p{0.7cm}||}
%    \hline
%    \space & \multicolumn{2}{l||}{\mbox{Similarity score}} & \multicolumn{2}{l||}{\mbox{XNLI accuracy}} \\ [0.5ex]\cline{2-5}
%     \space &\textbf{50K} & \textbf{100K} & \textbf{50K} & \textbf{100K} \\ 
%    \hline\hline
%    English (new tok.) & 0.725 &  0.786 &  55.7 & 59.4\\ \hline
%    Spanish order & 0.722 & 0.779 & 55.9 & 58.9 \\ [0.5ex] \hline
%    Greek order & 0.718 & 0.773 & 54.8 & 58.1 \\ [0.5ex] \hline
%    Chinese order & 0.723 & 0.774 & 55.1 & 58.1 \\ [0.5ex] \hline
%    Hebrew order & 0.725 & 0.781 & 55.2 & 59 \\ [0.5ex] \hline
%    Hindi order & 0.72 & 0.776 & 56.5 & 57.9 \\ [0.5ex] \hline
%    \end{tabular}
%    }

%    \caption{Results from the distillation process for the word reordering experiment.\vspace{-.3cm}}\label{tab:word-reorder}
%\end{table}

\begin{table}[h!]
    \centering
    \small{
    \begin{tabular}{|p{2.3cm}||p{0.8cm}|p{0.8cm}||p{0.75cm}|p{0.75cm}||}
    \hline
    \space & \multicolumn{2}{c||}{\mbox{ACS score}} & \multicolumn{2}{c||}{\mbox{XNLI accuracy}} \\ [0.5ex]\cline{2-5}
     \space &\textbf{50K} & \textbf{100K} & \textbf{50K} & \textbf{100K} \\ 
    \hline\hline
    English & 0.725 &  0.786 &  55.7 & 59.4\\ \hline
    Spanish order & -0.002 & -0.007 & +0.2 & -0.5 \\ [0.5ex] \hline
    Greek order & -0.007 & -0.013 & -0.9 & -1.3 \\ [0.5ex] \hline
    Chinese order & -0.002 & -0.012 & -0.6 & -1.3 \\ [0.5ex] \hline
    Hebrew order & -0.0 & -0.005 & -0.5 & -0.4 \\ [0.5ex] \hline
    Hindi order & -0.005 & -0.01 & +0.8 & -1.5 \\ [0.5ex] \hline
    \end{tabular}
    }

    \caption{Results from the distillation process for the word reordering experiment.\vspace{-.3cm}}\label{tab:word-reorder}
\end{table}

\vspace{-.2cm}
\subsection{Lexical Swapping}\label{sec:lexical}

We follow the steps described in Section \S\ref{sec:method-lexical} for Spanish, Greek and Simplified Chinese. When constructing the weighted bipartite graph, for Spanish and Greek we use the datasets \textit{Europarl+TED}, whereas for Simplified Chinese we use \textit{UM+TED}. Results are presented in Table \ref{tab:lexical-swap}. In this experiment, we observe a significant decrease in both the ACS score and the XNLI accuracy. The language that performs the worst is Simplified Chinese, with up to 0.092 degradation in the ACS score and up to 5.9\% in XNLI accuracy.

%\begin{table}[h!]
%    \centering
%    \small{
%    \begin{tabular}{|p{2.5cm}||p{0.7cm}|p{0.7cm}||p{0.7cm}|p{0.7cm}||}
%    \hline
%    \space & \multicolumn{2}{l||}{\mbox{Similarity score}} & \multicolumn{2}{l||}{\mbox{XNLI accuracy}} \\ [0.5ex]\cline{2-5}
%     \space &\textbf{50K} & \textbf{100K} & \textbf{50K} & \textbf{100K} \\ 
%    \hline\hline
%    English (new tok.) & 0.725 &  0.786 &  55.7 & 59.4\\ \hline
%    Spanish lexicon & 0.67 & 0.726 & 53.4 & 57.5 \\ [0.5ex] \hline
%    Greek lexicon & 0.652 & 0.713 & 51.6 & 56.1 \\ [0.5ex] \hline
%    Chinese lexicon & 0.646 & 0.694 & 50.9 & 53.5 \\ [0.5ex] \hline
%    \end{tabular}
%    }\caption{Results from the distillation process for the lexical swapping experiment.}\label{tab:lexical-swap}
%\end{table}

\begin{table}[h!]
    \centering
    \small{
    \begin{tabular}{|p{2.3cm}||p{0.8cm}|p{0.8cm}||p{0.75cm}|p{0.75cm}||}
    \hline
    \space & \multicolumn{2}{c||}{\mbox{ACS score}} & \multicolumn{2}{c||}{\mbox{XNLI accuracy}} \\ [0.5ex]\cline{2-5}
     \space &\textbf{50K} & \textbf{100K} & \textbf{50K} & \textbf{100K} \\ 
    \hline\hline
    English & 0.725 &  0.786 &  55.7 & 59.4\\ \hline
    Spanish lexicon & -0.055 & -0.06 & -2.3 & -1.9 \\ [0.5ex] \hline
    Greek lexicon & -0.073 & -0.073 & -3.4 & -3.3 \\ [0.5ex] \hline
    Chinese lexicon & -0.079 & -0.092 & -4.8 & -5.9 \\ [0.5ex] \hline
    \end{tabular}
    }\caption{Results from the distillation process for the lexical swapping experiment.\vspace{-.2cm}}\label{tab:lexical-swap}
\end{table}

These results suggest that variations in lexicons significantly impact the capacity to align a language with a pretrained representation space, thereby affecting transferability. To gain a deeper understanding of this phenomenon, we proceed to applying the same manipulation, this time selectively swapping only a subset of the words in the language.

\begin{figure}[h]
    \includegraphics[width=1.0\linewidth]{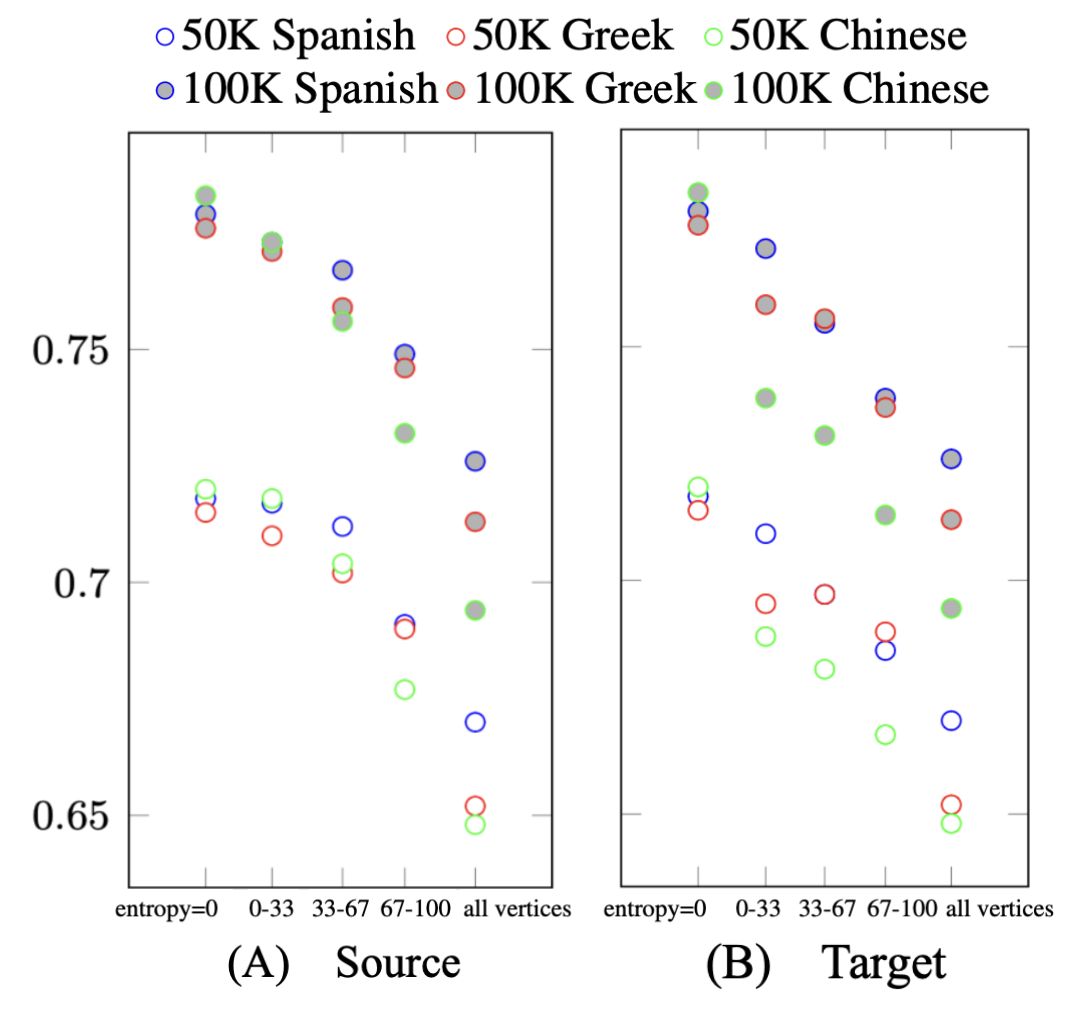}
    \caption{Results of filtering the \textbf{source} vertices in the graph in Figure (A) and results of filtering the \textbf{target} vertices in the graph in Figure (B). The horizontal axis represents the entropy values ranging from 0 to \textit{all vertices}. Numerical x values denote percentiles. The y axis represents ACS scores.}
    \label{fig:src-trg-entropy-filtering}
\end{figure}

\vspace{.1cm}\noindent\textbf{Entropy-based Lexical swapping}. In this experiment we filter the vertices of the bipartite graph based on their \textit{translation entropy} (see \S\ref{sec:method-lexical}) and then apply the lexical swapping manipulation. Figure \ref{fig:src-trg-entropy-filtering}{A} presents the outcome of filtering the source vertices, and Figure \ref{fig:src-trg-entropy-filtering}{B} shows the result of filtering the target vertices. In both cases, we split the set of vertices based on percentiles: into the ranges of 0-33, 33-67, and 67-100. In addition, we include an experiment where we exclusively swap words with zero entropy, and we add the results from the full lexical manipulation.

We observe a robust negative correlation between the entropy of the words we swap and the similarity scores. In all cases except for one (filtering Greek words of percentile 33-67), the higher the entropy of the words swapped, the worse the distillation process performs. Moreover, when we swap only 33\% of English word instances with low entropy, it has minimal impact on performance, but when we swap 33\% of word instances with the highest entropy, it results in a degradation of performance that is close to the degradation observed in the full lexical manipulation. We conclude that swapping in itself does not degrade performance; instead, most degradation results from the lexicons not being aligned in a one-to-one manner.

The absence of one-to-one alignment in the lexicons conceals two separate phenomena: synonymy and polysemy. In case of a synonymy, a specific word is translated to different words in different contexts, whereas in the case of polysemy, several distinct words are translated  to the same word. The first experiment (filtering the source words) mostly simulates the impact of synonymy, while the second experiment (filtering the target words) mostly simulates the impact of polysemy. Results imply that both phenomena have a substantial impact on cross-lingual transfer.

\section{Conclusion}

We leverage a knowledge distillation setup to explore the conditions that allow successful cross-lingual transfer. We apply various manipulations to English to alter specific language properties and assess their impact.

% option 1
%\textcolor{red}{First, we apply script substitution and observe no degradation in performance. Next, we examine the impact of word order. Unlike previous studies \cite{deshpande-etal-2022-bert, Chai:2022, wu2024oolong}, which made only subtle modifications to the constituent order in some experiments and inverted/shuffled all the words in others, we apply a manipulation that permutes many words in the sentence but still maintains a coherent syntactic structure. We believe that this manipulation provides us with a more controlled and nuanced understanding of the role of word order in cross-lingual transfer. Our initial observations imply that word order differences, if systematic, may not play a crucial role.}

% option 2
%\textcolor{blue}{First, we apply a script substitution and observe no degradation in performance. Next, we examine the impact of word order. Previous studies focusing on MLLMs \cite{deshpande-etal-2022-bert, Chai:2022, wu2024oolong} showed minor performance degradation when systematically adapting the word order of different languages, but a substantial decline when inverting or shuffling the entire sentence. Our findings within a knowledge distillation setup support these observations, suggesting that in this context as well, as long as the syntactic structure remains coherent, differences in word order between languages do not play a crucial role.}

First, we apply script substitution and observe almost no degradation in performance. Next, we examine the impact of word order. Unlike previous studies \cite{deshpande-etal-2022-bert, Chai:2022, wu2024oolong}, which made only subtle modifications to the constituent order in some experiments and inverted/shuffled all the words in others, we apply a manipulation that permutes many words in the sentence but still maintains a coherent syntactic structure. We believe that this manipulation provides us with a more nuanced understanding of how word order affects cross-lingual transfer. Our findings align with the survey by \citet{philippy-etal-2023-towards}, suggesting that as long as the syntactic structure remains coherent, the effect of word order is less substantial (compared to inversion/shuffling).

Finally, we swap words from the English lexicon with words from the target lexicon and observe a substantial degradation in performance. We use the notion of \textit{translation entropy} to explore the impact of swapping only a subset of words in the lexicon. This reveals that swapping the words with the highest entropy leads to a more substantial degradation in performance compared to words with lower entropy. These findings 
support our hypothesis: the more the lexicons align in a one-to-one manner, the better cross-lingual transfer will perform. 

To recap, Among the three manipulations we apply then, only lexical swapping was found to have a substantial effect. This suggests that when it comes to cross-lingual transfer, at least in the case of model distillation, lexicalization differences across languages may be more crucial than other linguistic factors such as word order. This finding offers valuable guidance for optimizing cross-lingual transfer systems.

\section*{Limitations}

Our work has several limitations (we intend to address them in future work). First, all our  experiments are conducted using a monolingual teacher model. We consider it important to examine the influence of multilingual pretraining. The potential impact of a representation space that is not tailored to a particular language could be substantial. Secondly, the sum of degradations resulting from the various manipulations we apply does not reach the degradation caused by cross-lingual transfer. This could stem from the fact that translations are not always accurate, but it can also indicate that we are missing an important determinant of cross-lingual transferability. Last, 
many other manipulations that impact the lexical semantics of the source languages were not considered here. For example, it would be 
valuable to apply the manipulation to a different subset of the language (e.g., enabling only synonymy but not polysemy, filtering by part-of-speech tags  etc.).

\section*{Acknowledgments}

We would like to express our gratitude to Ofir Arviv for his useful advice. 
This work was supported in part by the Azrieli Fellowship, the Vatat scholarship, and the Israel Science Foundation (grant no. 2424/21).

\bibliographystyle{acl_natbib}
\bibliography{anthology, custom}

\appendix

\section{Baseline Experiments $\&$ Degradation Analysis}\label{app:baseline- experiments}

\subsection{Training the student model on English}\label{app:english-english}

To understand how both the size of the data and the selected model architecture influence the quality of alignment we begin by training the student models on English. We train models of various architectures on subsets of various sizes from \textit{CC-100}. The tokenizer we use is the original tokenizer of the teacher model. Table \ref{tab:english-english-1} reports the average cosine similarity (ACS) of all the sentences in the separated test set when they are encoded once using the teacher model and once using the student model. Table \ref{tab:english-english-2} reports the accuracy on the XNLI test set in a zero-shot setting (the MLP built upon the teacher model achieved an accuracy of 71.2\%).

\begin{table}[h!]
    \begin{subtable}{0.45\textwidth}
    \small{
    \begin{tabular}{|p{2.5cm}||p{0.7cm}|p{0.7cm}|p{0.7cm}|p{0.7cm}||}
    \hline
      \space &\textbf{50K} & \textbf{100K} &\textbf{200K} & \textbf{1M} \\ \hline\hline
    3 hidden layers and 4 attention heads & 0.658 & 0.727 & 0.793 & 0.866\\ \hline 
    6 hidden layers and 6 attention heads &  \textbf{0.684} &  \textbf{0.754} &  \textbf{0.827} & 0.9\\ \hline
    9 hidden layers and 8 attention heads & 0.682 & 0.737 & 0.818 & \textbf{0.909}\\ \hline
    12 hidden layers and 12 attention heads & 0.677 & 0.74 & 0.81 & 0.901\\ [1ex] 
    \hline
    \end{tabular}
    }
    \caption{Average cosine similarity (ACS) of all the sentences paired with themselves in a held-out subset of \textit{CC-100}.}\label{tab:english-english-1}
    \end{subtable}\\

    \begin{subtable}{0.45\textwidth}
    \small{
    \begin{tabular}{|p{2.5cm}||p{0.7cm}|p{0.7cm}|p{0.7cm}|p{0.7cm}||}
    \hline
      \space &\textbf{50K} & \textbf{100K} &\textbf{200K} & \textbf{1M} \\ \hline\hline
    3 hidden layers and 4 attention heads & 53.7 & 57.6 & 61.3 & 63.3\\ \hline 
    6 hidden layers and 6 attention heads & 55.7 & \textbf{59.6} & \textbf{62.9} & 65.7\\ \hline
    9 hidden layers and 8 attention heads & \textbf{56.2} & 58.1 & 62.9 & \textbf{65.8}\\ \hline
    12 hidden layers and 12 attention heads & 54.3 & 58.1 & 61.8 & 64.9\\ [1ex] 
    \hline
    \end{tabular}
    }
    \caption{XNLI test accuracy in a zero-shot setting.}\label{tab:english-english-2}
    \end{subtable}
    \caption{Results from the distillation process with English as the target language for various architectures and various dataset sizes.}
\end{table}

Several conclusions can be drawn. First, we observe a robust correlation (Pearson correlation of 0.988) between the ACS score and zero-shot performance (the intrinsic and extrinsic performance respectively). This proves that the quality of the alignment with the pretrained representation space can be a useful tool for predicting zero-shot performance. Secondly, the results demonstrate that the size of the corpus has a great effect on the quality of the alignment. With 1M sentences, one can already train a student model that achieves an ACS score of 0.909 out of 1. Lastly, results indicate that the architecture of the student model has a relatively minor impact on performance. However, beyond a certain model size, training results reflect overfitting. As our main concern is low-resource languages, we decide to stick with the architecture that shows optimal performance in limited data scenarios: 6 hidden layers and 6 attention heads.

\subsection{Cross-lingual Transfer}\label{app:cross-lingual}

We conduct a cross-lingual experiment using the \textit{TED-2020} parallel corpus. Results are presented in Table \ref{tab:cross-lingual}. We present the outcomes of training the English encoders with both the original teacher's model tokenizer and a newly trained tokenizer (see Appendix \ref{app:tokenizers}). 

\begin{table}[h!]
    \centering
    \small{
    \begin{tabular}{|p{2.3cm}||p{0.75cm}|p{0.75cm}||p{0.75cm}|p{0.75cm}||}
    \hline
    \space & \multicolumn{2}{c||}{\mbox{ACS score}} & \multicolumn{2}{c||}{\mbox{XNLI accuracy}} \\ [1ex]\cline{2-5}
     \space &\textbf{50K} & \textbf{100K} & \textbf{50K} & \textbf{100K} \\ 
    \hline\hline
    English - teachers' tokenizer & 0.74 & 0.804 & 56.6 & 60.5\\ \hline 
    English - new \textit{CC-100} tokenizer & 0.725 &  0.786 &  55.7 & 59.4\\ \hline
    Spanish & 0.601 & 0.657 & 49.4 & 54\\ [1.5ex] \hline
    Greek & 0.574 & 0.632 & 49.9 & 53.1\\ [1.5ex] \hline
    Chinese & 0.555 & 0.6 & 40.9  & 46.5 \\ [1.5ex] \hline
    Hebrew & 0.545 & 0.606 & X & X \\ [1.5ex] 
    \hline
    \end{tabular}
    }

    \caption{Results from the distillation process (average cosine similarity scores and XNLI accuracies) for various languages using the \textit{TED-2020} parallel corpus.}\label{tab:cross-lingual}
\end{table}

We can see that the tokenizer's substitution results in only a minor performance degradation (0.015 points in ACS score when trained with 50K sentences), while the transition to a different language leads to a substantial decrease (0.124 when trained with 50K Spanish sentences). Unsurprisingly, languages closer to English in terms of phylogenetic distance, produce higher ACS scores and better zero-shot performance.

%%%%%%%%%%%%%%%%%%%%%%%%%%%%%%
\section{Word Reordering Algorithm}\label{app:reorder-algorithm}

We hereby describe the word reordering algorithm developed \citet{Arviv:2023}, that we apply to permute the words of the source sentences so that it will conform to the syntactic structure of the target language.
The algorithm relies on the statistics of the \href{https://universaldependencies.org/}{Universal Dependencies (UD)} treebank to permute the words of a sentence in one language so that they mimic the syntactic structure of another. 
The algorithm is built on the assumption that a contiguous subsequence, which constitutes a grammatical unit in the original sentence, should remain a contiguous subsequence after reordering, although the order of words within that subsequence may change. It operates, therefore, on a UD dependency tree, recursively permuting each sub-tree so that it will conform to the order of an equivalent sub-tree in the target language. 

Within each sub-tree, the reordering is applied based on the notion of \emph{pairwise ordering distributions}. Given a sentence $t$ in a language $L_t$ and its UD parse tree $T(t)$, which contains the set of dependency labels $\pi=(\pi_1,...,\pi_n)$, Arviv et al. denote the \emph{pairwise ordering distribution} in language $L_t$ of two UD nodes with dependency labels $\pi_i, \pi_j$, in a sub-tree with the root label $\pi_k$ by:
\begin{equation}
P_{\pi_k,\pi_i,\pi_j}=p;p\in[0,1]
\label{eq:ud_pairwise_dist}
\end{equation}
where $p$ stands for the probability of a node with a dependency label $\pi_i$ to be linearly ordered before a node with a label $\pi_j$, in a sub-tree with a root of label $\pi_k$, in a language $L_t$.\footnote{Note that a single node can act both as a representative of its sub-tree and the head of that sub-tree.} 

Given a sub-tree $T_i\in T(t)$, for each of its node pairs, these probabilities are formulated as a constraint:

\begin{equation}
    \pi_k:(\pi_i<\pi_j) = 
    \begin{cases}
        \mathbf{1} &\text{ if $P_{\pi_k,\pi_i,\pi_j}\geq0.5$} \\
        \mathbf{0} &\text{ otherwise}
    \end{cases}
\end{equation}

where $\pi_k:(\pi_i<\pi_j)=\mathbf{1}$ indicates that a node with label $\pi_i$ should be linearly ordered before a node with label $\pi_j$ if they are direct children of a node with label $\pi_k$. A constraint is said to be satisfied if and only if the node with label $\pi_i$ is indeed positioned in the sentence before the node with label $\pi_j$. For each individual sub-tree $T_i$, all its pairwise constrains are extracted, and an SMT solver is used to compute a legal ordering which satisfies all the constraints.\footnote{If it is not possible to fulfill all the constraints, the algorithm maintains the original order of the sub-tree.}

%%%%%%%%%%%%%%%%%%%%%%%%%%%%%%%
%%%%%%%%%%%%%%%%%%%%%%%%%%%%%%%
%%%%%%%%%%%%%%%%%%%%%%%%%%%%%%%

\section{Lexical manipulation: Implementation Details}\label{app:lexical-implementation-details}

\textbf{Tokenization and Lemmatization}. Before we perform word-to-word alignment, we have to separate the sentences' tokens and lemmatize them. For this purpose we use \href{https://trankit.readthedocs.io/en/latest/}{\textit{Trankit}} \cite{nguyen2021trankit}, a multilingual NLP toolkit based on \textit{XLM-R}. For Simplified Chinese, however, we prefer the \href{https://github.com/fxsjy/jieba}{Jieba} tokenizer.

\vspace{.1cm}\noindent\textbf{Graph Filtering}\label{par:graph-filtering}. When considering the filtering of the graph, we face two choices: we can either apply identical parameters for all languages or customize parameters for each language in a way that ensures a similar percentage of alignment instances is filtered from the graph. The first option maintains a similar level of noise across languages but has a drawback: when we apply the lexical manipulation, removing a high percentage of alignment instances from the graph results in selecting the most common word too frequently (see step \ref{item:most-common-word} in the lexical manipulation procedure), and therefore loses the ability to make meaningful comparisons across different languages.

In our chosen method, we aim for the middle ground. We start by removing from the graph every edge with a weight below the threshold of 5 to exclude matches that are not substantial. Then, for each language, we set a specific threshold to remove edges whose weight is relatively small compared to other edges originating from the same vertex. We set this second threshold in such a way that for each language, a total of approximately 12\% of the alignment instances are filtered out. In the case of Spanish and Greek, the appropriate threshold is 2\%, while for Simplified Chinese, it is 0.15\%.

%%%%%%%%%%%%%%%%%%%%%%%%%%%%%%
\section{Comparing Alignments to Gold Standard}\label{app:golden-alignments}
We evaluate the alignment results of our algorithm against the original Simalign alignments, using the gold standard provided by \cite{Graca:2008}. We focus our comparison on the \textit{English--Spanish} alignments, as this language pair is the sole one utilized in our research. The obtained results are as presented in Table \ref{tab:alignments-comparison}. We can see that our algorithm's outputs achieve higher precision but lower recall, resulting in a slightly better F1 score overall. 

To further understand this point, let us examine a specific example (as others are similar): the sentence \textit{We take note of your statement} is translated into \textit{Tomamos nota de esa declaración}. While the Simalign auto-aligner aligns \textit{your} with \textit{esa}, our algorithm filters out this alignment, as these two words are rarely translations of one another in the larger corpus. Although we miss a correct alignment in the gold standard, this approach conforms to our goal of mapping lexicons consistently.

\begin{table}[h!]
    \small{
    \begin{tabular}{|p{2.5cm}||p{1.1cm}|p{1.1cm}|p{1cm}||}
    \hline
      \space &\textbf{Precision} & \textbf{Recall} &\textbf{F1} \\ \hline\hline
    Original simalign & 73.39 & \textbf{90.8} & 81.17 \\ \hline 
    Our algorithm &  \textbf{76.55} &  86.58 &  \textbf{81.26} \\ \hline
    \end{tabular}
    }
    \caption{Comparison of Alignments to Gold Standard}\label{tab:alignments-comparison}
\end{table}

%%%%%%%%%%%%%%%%%%%%%%%%%%%%%%

\section{Datasets} \label{app:datasets}

\vspace{.1cm}\noindent\textbf{TED}. The bitext we use for training and intrinsic evaluation of the student models throughout all manipulation experiments is the \href{https://opus.nlpl.eu/TED2020.php}{\textit{TED-2020}} parallel corpus  \citep{Reimers:2020}. This corpus contains a crawl of nearly 4000 TED transcripts from July 2020, which have been translated into over 100 languages by a global community of volunteers. We select this corpus because it contains languages from different language families, and because its translations are of relatively high quality. To further simplify it, we lowercase the entire dataset and filter it to include only sentences with familiar characters, up to one punctuation mark, and word counts ranging from 4 to 16.\footnote{Except for Simplified Chinese, where, due to the different nature of logographic writing systems, we filter by counting 5-25 symbols.} The corpus is licensed under CC BY–NC–ND 4.0.

\vspace{.1cm}\noindent\textbf{CC-100}. When we require a larger corpus, but not necessarily a parallel one, we turn to the \href{https://huggingface.co/datasets/cc100}{\textit{CC-100}} corpus \citep{Weznek:2020}. This corpus serves us for training the tokenizers (See Appendix \ref{app:tokenizers}). Our goal in tokenizer training is to prevent over-specialization on the limited number of sentences used for training the student model. To achieve this, we use the largest and most diverse corpus available to us, namely, CC-100. Our experiments simulate scenarios where there is a significant amount of monolingual data for a specific language but minimal parallel data, which is the case for many low-resource languages. Additionally, we employ this corpus for our \textit{English--English} experiments (see Appendix \ref{app:english-english}). In both cases, we apply the same simplifying process as for TED, bringing the formats of the two datasets closer to each other. The license of this corpus is unspecified by the authors, but they state that they will make it publicly available.

\vspace{.1cm}\noindent\textbf{XNLI}. For extrinsic evaluation we use Natural Language Inference (NLI), as it is a well-known   sentence level semantic task. The task is to determine the inference relation between two sentences: \textit{entailment}, \textit{contradiction}, or \textit{neutral}. The corpus we use is the \href{https://github.com/facebookresearch/XNLI}{\textit{Cross-lingual Natural Language Inference (XNLI)}} \cite{conneau:2018}, which contains 15 different languages. There is no need to apply a simplifying process to this dataset, as the sentences are already relatively short and do not contain unconventional characters. The corpus is licensed under CC BY-NC 4.0.

\vspace{.1cm}\noindent\textbf{Europarl}. In order to extract a bipartite graph which is statistically meaningful for our lexical manipulation, we require a large parallel corpus. We use \href{https://www.statmt.org/europarl/}{\textit{Europarl}}, which consists of the proceedings of the European Parliament from 1996 to 2012. This corpus contains only European languages, so we must turn to other sources when experimenting with languages from different language groups. The corpus is freely available.

\vspace{.1cm}\noindent\textbf{UM}. We extract our \textit{English--Chinese} bipartite graph from the \href{http://nlp2ct.cis.umac.mo/um-corpus/}{\textit{UM}} parallel corpus \citep{Tian:2014}. It contains more than 2M \textit{English--Chinese} sentence pairs from a great variety of domains. The corpus is licensed under CC BY–NC–ND 4.0.

%%%%%%%%%%%%%%%%%%%%%%%%%%%%%%%
%%%%%%%%%%%%%%%%%%%%%%%%%%%%%%%
%%%%%%%%%%%%%%%%%%%%%%%%%%%%%%%

\section{Tokenizers}\label{app:tokenizers}

When training English models, we examine two different tokenizers: the original teachers' tokenizer, and a new tokenizer we train on the simplified \textit{CC-100} corpus. In the case of other languages, we train a new tokenizer on the simplified \textit{CC-100} corpus. 

The cases of the script and lexical manipulations each require its special treatment. In the case of the script manipulations, we create an artificial language which is composed of English words with foreign symbols, so we require a tokenizer which is familiar with this specific language. We simply apply the manipulation to the English \textit{CC-100} corpus and train a tokenizer on the transformed sentences.

In the case of the lexical manipulation, we swap some English words while retaining others, resulting in an artificial language which is a fusion of two languages. Therefore, a bilingual tokenizer is required. We train a bilingual tokenizer for each language pair using the \textit{CC-100} corpus.\footnote{Note that the word-reorder manipulation, as it maintains the same set of words as in the original sentence, does not require any special treatment.}

%%%%%%%%%%%%%%%%%%%%%%%%%%%%%%

\clearpage\end{CJK*}
\end{document}